\documentclass[runningheads]{llncs}

\usepackage[T1]{fontenc}
\def\doi#1{\href{https://doi.org/\detokenize{#1}}{\url{https://doi.org/\detokenize{#1}}}}
\usepackage{graphicx}
\usepackage{listings}
\usepackage{graphicx}
\usepackage{subcaption}
\usepackage{xcolor}
\usepackage{cite}
\usepackage{xspace} 

\usepackage{amsmath}
\usepackage{amsfonts}
\usepackage[belowskip=-12pt,aboveskip=0pt]{caption}

\def\PCADeepSSM{PCA-DeepSSM\@\cite{DeepSSM}\xspace}
\def\PPCADeepSSM{PPCA-DeepSSM\@\cite{uncertaindeepssm}\xspace}
\def\PPCAODeepSSM{PPCA-Offset-DeepSSM\@\cite{Parisot2018}\xspace}
\def\VIBDeepSSM{VIB-DeepSSM\@\xspace}

\newcommand{\etal}{\textit{et al.}~}
\newcommand{\ie}{i.e.,~}

\newcommand{\eg}{e.g.,~}

\newcommand{\x}{\mathbf{x}}
\newcommand{\y}{\mathbf{y}}
\newcommand{\z}{\mathbf{z}}

\newcommand{\mcX}{\mathcal{X}}
\newcommand{\mcY}{\mathcal{Y}}
\newcommand{\mcZ}{\mathcal{Z}}

\renewcommand{\a}{\mathbf{a}}

\newcommand{\xn}{\x_n}
\newcommand{\yn}{\y_n}
\newcommand{\zn}{\z_n}
\newcommand{\an}{\a_n}

\renewcommand{\xi}{\x_i}

\newcommand{\R}[1]{\mathbb{R}^{#1}}

\newcommand{\U}{\mathbf{U}}

\newcommand{\bmu}{\boldsymbol{\mu}}

\renewcommand{\L}{\mathcal{L}}

\newcommand{\N}{\mathcal{N}}

\newcommand{\KL}[2]{\operatorname{KL}\left[ #1 \| #2 \right]}
\newcommand{\E}[2]{\mathbb{E}_{#1}\left[#2\right]}

\begin{document}

\title{From Images to Probabilistic Anatomical Shapes: A Deep Variational Bottleneck Approach \thanks{Provisionally accepted for MICCAI 2022.}}

\titlerunning{From Images to Probabilistic Anatomical Shapes}

\author{Jadie Adams\inst{1,2}  \and
Shireen Elhabian\inst{1,2}}
\authorrunning{Adams and Elhabian}

\institute{Scientific Computing and Imaging Institute, University of Utah, UT, USA \and
School of Computing, University of Utah, UT, USA \\
\email{ \{jadie, shireen\}@sci.utah.edu }
}

\maketitle              
\vspace{-0.15in}
\begin{abstract}
Statistical shape modeling (SSM) directly from 3D medical images is an underutilized tool for detecting pathology, diagnosing disease, and conducting population-level morphology analysis. Deep learning frameworks have increased the feasibility of adopting SSM in medical practice by reducing the expert-driven manual and computational overhead in traditional SSM workflows. However, translating such frameworks to clinical practice requires calibrated uncertainty measures as neural networks can produce over-confident predictions that cannot be trusted in sensitive clinical decision-making. Existing techniques for predicting shape with aleatoric (data-dependent) uncertainty utilize a principal component analysis (PCA) based shape representation computed in isolation of the model training. This constraint restricts the learning task to solely estimating pre-defined shape descriptors from 3D images and imposes a linear relationship between this shape representation and the output (i.e., shape) space. In this paper, we propose a principled framework based on the variational information bottleneck theory to relax these assumptions while predicting probabilistic shapes of anatomy directly from images without supervised encoding of shape descriptors. Here, the latent representation is learned in the context of the learning task, resulting in a more scalable, flexible model that better captures data non-linearity. Additionally, this model is self-regularized and generalizes better given limited training data. Our experiments demonstrate that the proposed method provides an accuracy improvement and better calibrated aleatoric uncertainty estimates than state-of-the-art methods. 

\end{abstract}

\section{Introduction}
Statistical shape modeling (SSM) is an enabling tool in medicine and biology to quantify population-specific anatomical shape variation.  
SSM can detect pathological morphologies tied to impaired function and answer clinical hypotheses regarding  anatomical cohorts (e.g., \cite{bhalodia2020quantifying, harris2013cam, atkins2017quantitative}).
Two effective representations of shape in SSM are deformation fields and \textit{landmarks} - the former captures implicit transformations between images/shapes and a pre-defined atlas, 
while the latter are sets of explicit points consistently defined on the shape surface such that they are in correspondence across the population \cite{RTW:Tho17, sarkalkan2014statistical}.
While the framework we propose is agnostic to the choice of shape representation, we demonstrate the approach using landmarks, both to be consistent with existing relevant methods and because landmarks are more intuitive and interpretable for statistical analyses and visualizing results \cite{zachow2015computational,sarkalkan2014statistical}.
Existing computational workflows, such as ShapeWorks \cite{cates2007shape, cates2017shapeworks}, automatically place dense sets of landmarks or \textit{correspondence points} on shapes segmented from 3D medical images.
However, generating such a set of correspondence points or \textit{point distribution model} (PDM) requires expert-driven, data-intensive steps: segmenting the anatomy of interest from 3D images, data preprocessing, shape registration, and correspondence optimization along with hyperparameter tuning.

Deep learning has alleviated these burdens by providing end-to-end solutions for mapping unsegmented 3D images (i.e., CT or MRI) to PDM with little preprocessing \cite{DeepSSM, bhalodia2021deepssm,uncertaindeepssm,Parisot2018}. 
While the traditional SSM pipelines take hours, trained networks can discover statistical representations of anatomies directly from new images in seconds, creating the potential to streamline the adoption of SSM in research and practice.
DeepSSM \cite{DeepSSM, bhalodia2021deepssm} is one such state-of-the-art framework that provides SSM estimates that perform statistically similar to traditional methods in downstream tasks \cite{deepssm-afib}. 
DeepSSM relies on a supervised 
latent representation computed using principal component analysis (PCA) in advance of model training to enable data augmentation and incorporate prior shape knowledge, as has been done in related image-based tasks (\eg \cite{DeepSSM,huang2017heartnet,milletari2017stats,xie2017deepshape,zheng2015detection}).
However, PCA supervision imposes a linear relationship between the latent and the output space and restricts the learning task to strictly SSM prediction. Additionally, PCA does not scale in the case of large sets of high-dimensional shape data.

Another caveat of DeepSSM, and deep learning models at large, is that they can produce overconfident estimates that can not be blindly assumed to be accurate, especially in sensitive decision-making settings such as clinical practice.
There are two forms of uncertainty in such frameworks; aleatoric (or data-dependent) and epistemic (or model-dependent uncertainty, which can be explained away given enough training data) \cite{Kendall2017}. 
Learning from medical imaging data poses a challenge because scans can vary widely in quality and often suffer from coarse resolution, artifacts, and noise.
These factors signify a strong need to capture uncertainty inherent in the input data; thus here, we chose to focus on aleatoric uncertainty quantification.
Progress has been made in inferring probabilistic SSM from images to capture aleatoric uncertainty measures \cite{uncertaindeepssm, Parisot2018}.
However, existing probabilistic SSM models share the same limitations as DeepSSM in that they rely on a predetermined PCA-based latent representation.

Here, we present a novel formulation for inferring statistical representations of anatomies from unsegmented images that mitigates these limitations by learning a latent representation in tandem via a variational information bottleneck \cite{deepvib}. 
In information bottleneck (IB) theory, a stochastic encoding $\z$ captures the minimal sufficient statistics required of input $\x$ to predict the output $\y$ \cite{tishbyIB}.
The encoding $\z$  and model parameters $\Theta$ are estimated by maximizing the IB objective: 
\begin{equation}\label{eqn:IB}
    \operatorname{argmax}_\Theta I(\y,\z; \Theta) - \beta I(\x,\z; \Theta) 
\end{equation}
where $I$ is the mutual information and $\beta$, is a Lagrangian multiplier that controls the trade-off between predictive accuracy (encouraging $\z$ to be maximally expressive about $\y$) and model complexity (encouraging $\z$ to be maximally compressive about $\x$). 
By leveraging this theory to learn a latent representation in our proposed model, we achieve better accuracy and uncertainty calibration than existing state-of-the-art techniques.
The advantages of this formulation include the following.
\begin{itemize}
    \item The latent representation is learned in the context of the learning task rather than in isolation, resulting in better accuracy and a more scalable framework. 
    \item The proposed model better captures the true shape distribution by allowing for a non-linear relationship between the latent encoding and output space. 
    \item This formulation is self-regularized and thus generalizes better under limited training data without ad-hoc regularization methods.
    \item Because the latent space is unsupervised, the proposed model is general enough to accommodate other learning tasks. For example, it could be used to predict SSM plus additional clinically relevant quantities of interest.
    \item The proposed model outperforms existing state-of-the-art methods in both predictive accuracy and uncertainty calibration.
\end{itemize}

\section{Methods}

Given an unsegmented image $\xn \in \R{H \times W \times D}$, the task is to predict a statistical representation of an anatomy-of-interest in the form of a PDM $\yn \in \R{3M}$, which is a set of $M$ 3D correspondence points, and the associated point-wise estimates of aleatoric uncertainty $\an \in \R{3M}$. 
This task is solved by training a model with parameters $\Theta$ using a population of $N-$unsegmented 3D images $\mcX = \{\xn \}_{n=1}^{N}$ and their corresponding PDMs denoted $\mcY = \{\yn \}_{n=1}^{N}$.
The proposed method and state-of-the-art models used in comparison utilize a bottleneck and latent encoding $\mcZ = \{\zn \}_{n=1}^{N}$ where $\zn \in \R{L}$ and $L \ll 3M$.

\subsection{VIB-DeepSSM Formulation} 

\begin{figure}[ht]
\begin{center}
    \includegraphics[width=\textwidth]{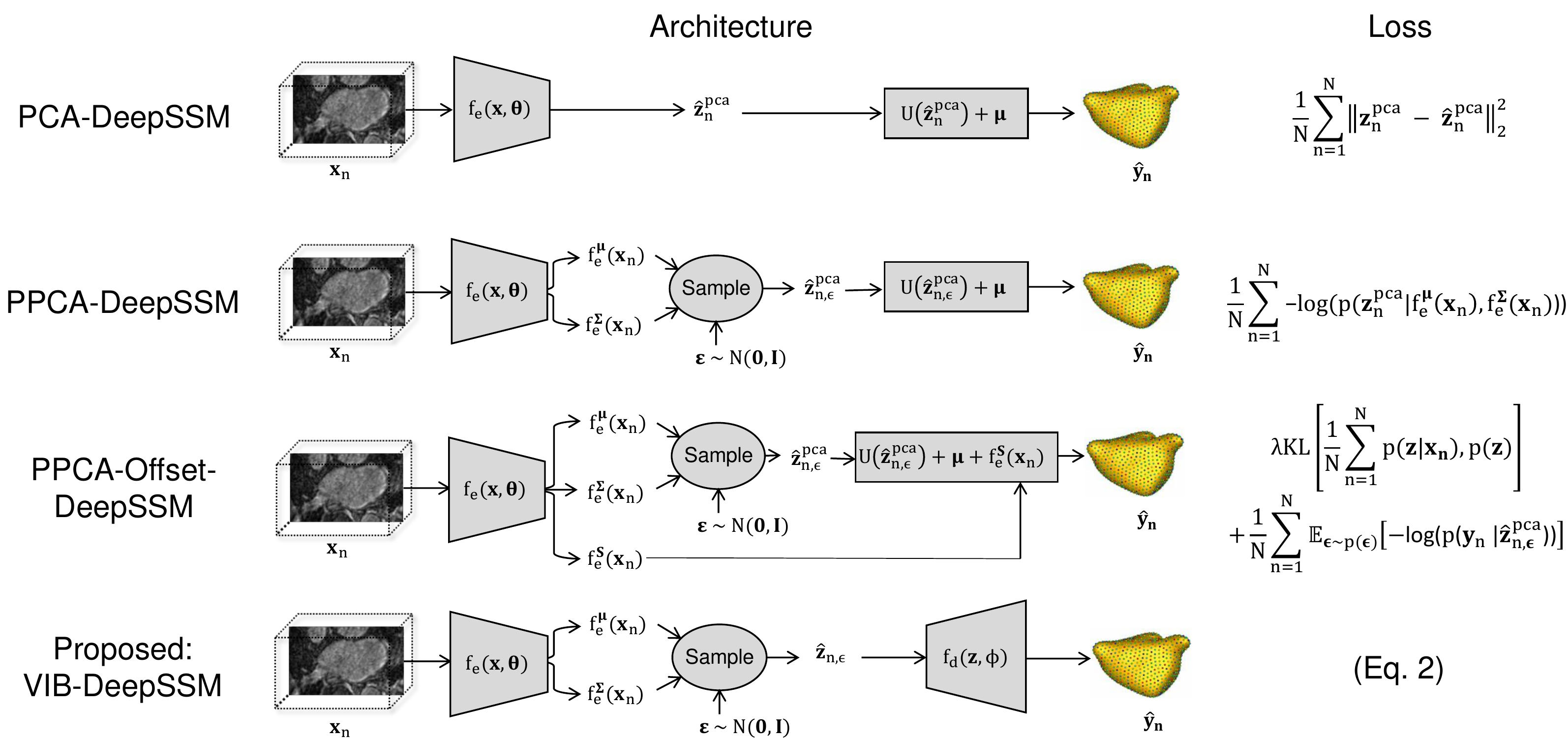}
    \caption{\textbf{Model Variants:} Architecture and loss of the proposed method and baseline models in comparison are shown. In each case, the encoder $f_e^\mu(\x;\theta)$ has the same architecture as in DeepSSM \cite{DeepSSM}; five convolutional layers with batch normalization followed by two fully connected layers. Only the output size of the last layer of the encoder is variant-dependent. In \VIBDeepSSM, the decoder $f_d(\z; \phi)$ is comprised of three fully connected layers with non-linear activations. }
    \label{fig:architecture}
\end{center}
\end{figure}

In the proposed method, denoted \textbf{\VIBDeepSSM}, the latent encoding is learned by minimizing the IB objective (Eq. \ref{eqn:IB}).
Given that directly calculating mutual information is ill-posed in this context, variational inference provides a way to approximate the problem given an empirical data distribution.
In "Deep Variational Information Bottleneck"\cite{deepvib}, the IB model is parameterized via a neural network by minimizing the derived theoretical lower bound on the IB objective:  

\begin{equation}\label{eqn:vib-loss}
    \L_{VIB} = \frac{1}{N} \sum_{n=1}^N \E{\epsilon \sim p(\epsilon)}{-\log{ q_\phi(\y_n | \z_{n,\epsilon})}} + \beta \KL{p_\theta(\zn|\xn)}{r(\z)}
\end{equation}
This loss balances a Kullback-Leibler divergence (KL) and a data fidelity (or reconstruction) term formulated as negative log-likelihood (NLL) with the Lagrangian multiplier $\beta \in [0,1]$ to learn model parameters $\Theta = \{\phi, \theta\}$.
The KL term encourages maximal compression of $\x$, while the reconstruction term encourages maximal expression of $\y$. 
We define a stochastic encoder of the form $p_\theta(\z|\x) = \mathcal{N}(\z|f_e^\mu(\x;\theta),f_e^\Sigma(\x; \theta))$, where $f_e$ is a convolutional network parameterized by $\theta$ that maps input image to a Gaussian latent distribution. 
Here, $r(\z)$ is a variational approximation to the marginal distribution of $\z$, which we  select to be a fixed $L$-dimensional spherical unit Gaussian, $r(\z) = \N(\z|\mathbf{0},\mathbb{I})$. 
The decoder, $q_\phi(\y|\z)$, serves as a variational approximation to the intractable $p(\y|\z)$. 
We define it as $q_\phi(\y|\z) = f_d(\z; \phi)$, where $f_d$ is an MLP parameterized by the weights $\phi$ (with non-linear activation functions) that maps the latent representation to correspondence points (Fig. \ref{fig:architecture}).
To enable gradient calculation, posterior samples $\z_{n,\epsilon}$ are acquired using the reparameterization trick \cite{kingma2014autoencoding}: $\z_{n,\epsilon} = f_e^\mu(\x_n;\theta) + \epsilon f_e^\Sigma(\x_n; \theta)$ where $\epsilon \sim \mathcal{N}(\mathbf{0},\mathbb{I})$. 
In our experiments, the expectation is estimated using 30 samples. 
The uncertainty is quantified as the entropy of this distribution $p(\y|\z)$.

\vspace{-0.1in}

\vspace{-0.1in}
\subsection{Baseline Models in Comparison}

We compare the proposed model against the state-of-the-art deterministic formulation that does not predict uncertainty, \PCADeepSSM, as well as two state-of-the-art stochastic formulations that do predict uncertainty: \PPCADeepSSM and \PPCAODeepSSM (Fig. \ref{fig:architecture}). 
In these frameworks, the latent representation is supervised via PCA scores computed from the training correspondence points $\yn$. 
The latent space dimension, $L$, is chosen such that $95\%$ of variability is preserved. 
We denote the PCA scores $\mcZ^{pca} = \{\zn^{pca} \}_{n=1}^{N}$ where $\zn^{pca} \in \R{L}$.
Thus $\yn \approx \U\zn^{pca}+\bmu$ where $\U \in \R{3M \times L}$ is the matrix of eigenvectors and $\bmu \in \R{3M}$ is the mean of the training correspondence points. 
In the proposed model, \VIBDeepSSM, the latent space is not supervised, yet its dimensionality should be predefined.
We found that PCA provides a reasonable, reproducible way to estimate latent dimension size in the unsupervised case as well, thus in \VIBDeepSSM, we use the same $L$-value, though this is not a requirement. 

\textbf{\PCADeepSSM} is a deterministic framework that uses PCA scores as a supervised latent space. The decoder is a single \textbf{fixed}, fully connected layer without activation with the PCA basis as weights and mean shape as bias. The loss function is mean square error (MSE) between ground-truth and predicted PCA scores.

\textbf{\PPCADeepSSM} is equivalent to Uncertain-DeepSSM \cite{uncertaindeepssm}, except drop-out is removed as we are not analyzing epistemic uncertainty quantification. 
Here the latent space is supervised and the encoder is stochastic, where PCA variance serves as a regularization term and is learned implicitly via the loss (NLL of the true PCA scores given the predicted distribution). 
The decoder is fixed PCA reconstruction, as it is in \PCADeepSSM, but here the predicted correspondence points are acquired by averaging over posterior samples.

\textbf{\PPCAODeepSSM} is based on the approach in T{\'{o}}thov{\'{a}} \etal \cite{Parisot2018}, but here the architecture is altered to predict 3D PDM instead of 2D shape vertices. 
In this formulation, the encoder also predicts a distribution of PCA scores; however, in addition, an offset term $f_e^s(\x) \in \R{3M}$ is predicted and added to the point prediction (Fig \ref{fig:architecture}).
The offset term helps address the linearity restriction, but it is unregularized and thus prone to overshoot. 
An ad hoc regularization scheme is utilized in this formulation in the form of KL divergence between an assumed prior distribution of $\z$ (selected to be $p(\z) = \mathcal{N}(0,I)$) and the observed one in the training set. 
The loss is the PDM NLL plus this KL term (Fig \ref{fig:architecture}).
Note that in this loss, the sum of the $N$ $\z$ predicted distributions (i.e., the aggregate posterior) is regularized by the KL term, and in the VIB loss (Eq. \ref{eqn:vib-loss}), each of the $N$ $\z$ predicted distributions are individually regularized. Hence, the VIB loss effectively minimizes the KL divergence between the aggregate posterior and the latent prior while minimizing the mutual information between $\x$ and $\z$ \cite{hoffman2016elbo}. This behavior is dictated by the information bottleneck; learn the $\z$ descriptor that is minimally informative of the input image $\x$ but maximally predictive of the shape $\y$.

\section{Results}
\vspace{-0.05in}

To best analyze accuracy and uncertainty calibration across model variants, we select a large, highly variable dataset for experiments. 
In the original DeepSSM \cite{DeepSSM} formulation, a data augmentation scheme is used to create additional training samples. 
However, this technique relies on PCA, which we intend to remove as a requirement; thus here training is done without augmentation.

\subsection{Left Atrium Dataset}
\begin{figure}[ht]
    \includegraphics[width=\textwidth]{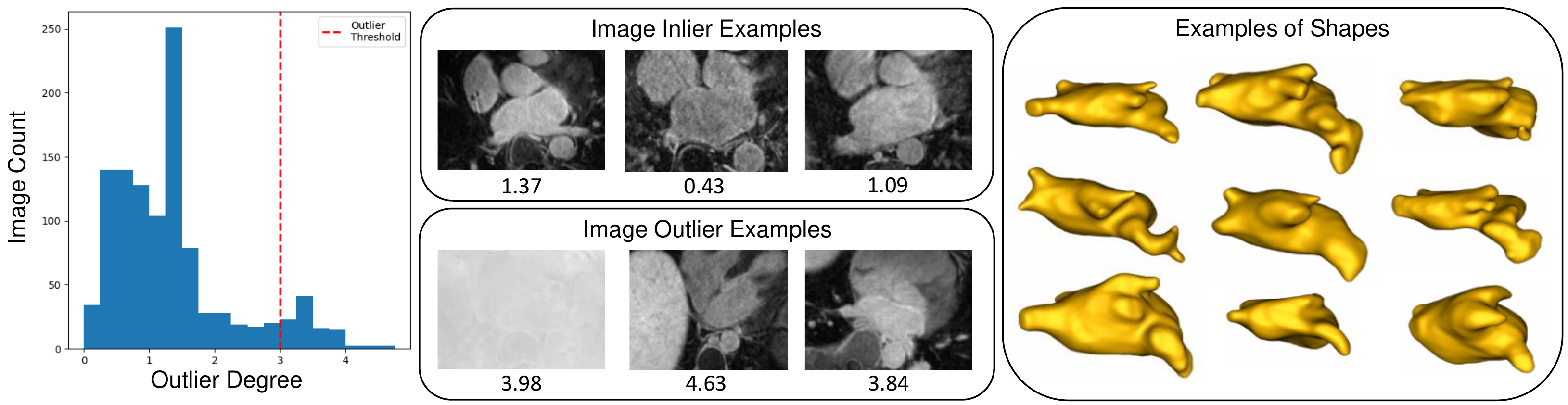}
    \caption{The distribution of image outlier degrees is displayed with examples of image slices and the corresponding outlier degree. Outlier images tend to be over-exposed with low contrast or contain artifacts. Examples of ground truth meshes are displayed from the top view to demonstrate the significant shape variability.} \label{fig:img_out}
\end{figure}

\vspace{-0.1in}
In experiments, we use a set of 1001 anonymized LGE MRI images of the left atrium (LA) from unique atrial fibrillation patients. 
The images were segmented by domain experts with spatial resolution $0.65 \times 0.65 \times 2.5$ mm$^3$ and the endocardium wall was used to cut off pulmonary veins. 
Significant morphological variations are expected to be in the left atrium appendage or LAA (varies immensely in size across patients), the pulmonary veins (vary in number and size across patients), and in the mitral valve (for which there are no defining image features for segmentation). 
This dataset is appropriate for aleatoric uncertainty analysis because, in addition to the large shape variation, the input images vary widely in intensity and quality, and LA boundaries are blurred and have low contrast with the surrounding structures (Fig. \ref{fig:img_out}). 
To enable calibration analysis, we define a specific testing set of images with high uncertainty using an outlier degree computed on input images that combines within- and off-subspace distances \cite{moghaddam1997probabilistic, uncertaindeepssm}.
By thresholding on outlier degree at 3 (Fig. \ref{fig:img_out}), we define an \textit{outlier test} set of size 78. 
We randomly split the remaining samples (90\%, 10\%, 10\%) to get a training set of 739, a validation set of 92, and an \textit{inlier test} set of 92. 

\begin{figure}[ht]
    \includegraphics[width=\textwidth]{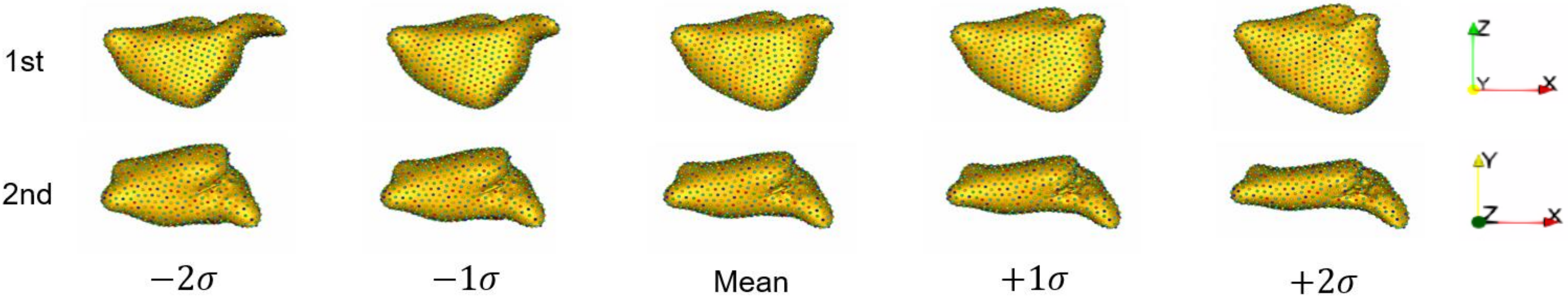}
    \caption{Modes of variation from the training PDM generated using ShapeWorks. The primary mode shown from anterior view captures the LAA elongation. The secondary mode shown from top view captures the LA sphericity.} \label{fig:modes}
\end{figure}

Acquiring target correspondence points for the training images requires generating a PDM via the traditional SSM pipeline (i.e., segmentation, preprocessing, and optimization). 
We employ the open-source \textit{ShapeWorks} software \cite{cates2017shapeworks} to align and crop the images and binary segmentations and optimize a PDM with 1024 landmarks per shape. 
The resulting input images are size (166, 120, 125).
Given the optimized training PDM, we generate target PDMs for the validation and test sets for analysis by optimizing landmarks positions on these samples with the training landmarks fixed so that the statistics of the validation and test samples are not reflected in the training PDM. 
The first two modes of variations in the training PDM are shown in Fig. \ref{fig:modes}.

\subsection{Training Scheme}
\VIBDeepSSM and the baseline methods are trained on 100\%, 80\%, 40\%, 20\%, and 10\% of the training data to analyze robustness in low-sample-size scenarios.
We select subsets of the training data using stratified random sampling by clustering the training PDMs and selecting data subsets so that each cluster is equally represented.
All variants are implemented in PyTorch and training is run on GTX1080Ti GPUs with Adam optimization \cite{adam}, a fixed learning rate of $5e^{-5},$ a batch size of 6, Xavier weight initialization \cite{xavier2010initialization}, and Parametric ReLU \cite{he2015rectifiers} activation. 
Models are trained until the validation PDM MSE has not decreased in 50 epochs. 
A loss burn-in scheme is used for stochastic variants to allow the model to learn to predict deterministic shapes before estimating variance. 
This is done using a linear combination of the given stochastic loss and deterministic loss (PDM MSE), where the weight changes over a given number of epochs from entirely deterministic to entirely stochastic.
We set the number of burn-in epochs to be proportional to the training data size (\ie 100 epochs for 100\%, 80 epochs for 80\%, etc.). 
To increase stability when training stochastic encoders, we predict the log of the latent variance, ensuring positivity and removing the potential for division by zero.
We empirically set hyperparameters using the validation MSE and find for \PPCAODeepSSM, $\lambda = 100$ (tested range: $[1e^{-4},1e^{4}]$) and for \VIBDeepSSM, $\beta = 0.01$ (tested range: $[1e^{-8},1]$).

\subsection{Accuracy Analysis}
\begin{figure}[ht]
    \begin{center}
    \includegraphics[width=\textwidth]{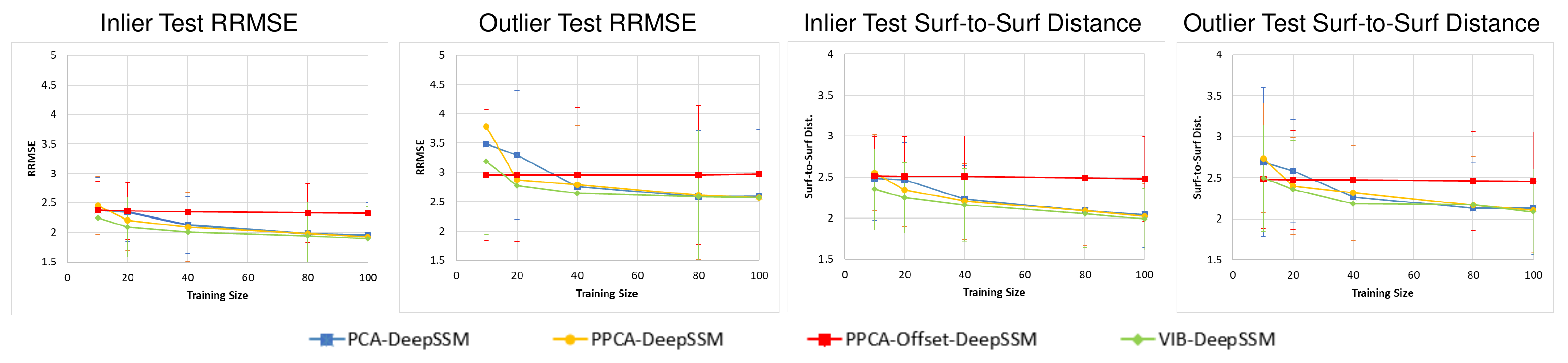}
    \caption{\textbf{Accuracy:} The mean accuracy is shown for each model variant with error bars representing standard deviation. Overall, \VIBDeepSSM performs best.} \label{fig:accuracy}
    \end{center}
\end{figure}

Prediction accuracy is evaluated by calculating the relative root mean square error (RRMSE) between the true and predicted correspondence points. 
This is evaluated sample-wise, \textbf{RRMSE} $= \sqrt{(||\yn - \hat{\y}_n||^2)/3M}$, and point-wise where for point $i$, \textbf{Point-RRMSE} $= \sqrt{(||\yn^i - \hat{\y}_n^i||^2)/3}$.
We also analyze prediction accuracy using the \textbf{surface-to-surface} distance between the ground truth mesh of the anatomy and one constructed from the predicted PDM.
To construct a mesh from predicted points, we find the closest example in the training set and apply the warp between the points to its mesh using \cite{wang2015linear}. 
The accuracy using different training sizes is summarized in Fig. \ref{fig:accuracy}. 
The proposed method has a similar or better accuracy on both testing inliers and outliers than the baseline methods given any training size.
Accuracy is most notably improved when training data is limited.
Note that for \PPCAODeepSSM, accuracy does not significantly improve with increased training data. 
This suggests that the model is over-regularized, however decreasing the $\lambda$ parameter results in unstable training.
This suggests \VIBDeepSSM provides a more principled way to capture data non-linearity and self-regularization.

\vspace{-0.1in}
\vspace{-0.2in}

\subsection{Uncertainty Calibration Analysis}
\begin{figure}[ht]
    \begin{center}
    \includegraphics[width=\textwidth]{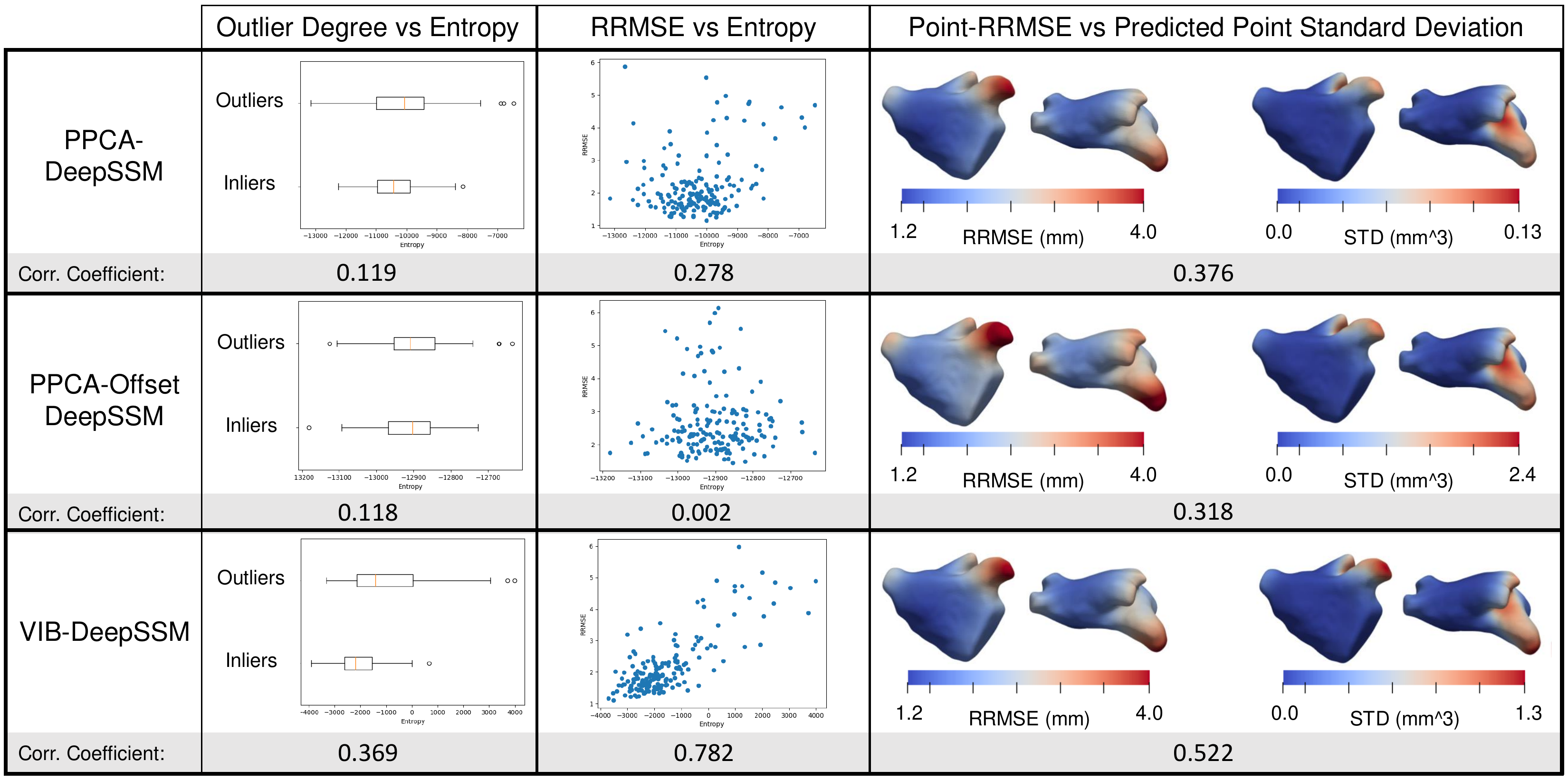}
    \caption{\textbf{Uncertainty Calibration} Plots and correlation coefficients show the uncertainty calibration of each model variant. Heat maps show average quantities on a representative mesh.} \label{fig:calibration}
    \end{center}
\end{figure}

To assess uncertainty calibration, we consider the correlation between (1) the input image outlier degree and entropy of $p(\y|\z)$, (2) RRMSE and entropy of $p(\y|\z)$, and (3) point-RRMSE and predicted point-wise standard deviation (volume of the 3D Gaussian for each landmark/point). 
We quantify correlation using the Pearson correlation coefficient, where a higher value suggests the model more effectively identifies input with high uncertainty and areas of low prediction confidence. 
Our models predicted uncertainty is better calibrated both in relation to outlier degree and error (Fig. \ref{fig:calibration}).
The heat maps generated via interpolation in Fig. \ref{fig:calibration} show that point uncertainty is correctly predicted to be higher in these areas where the error is highest (LAA, pulmonary veins, and mitral valve).

\vspace{-0.1in}

\section{Conclusion}
We presented a novel principled framework based on the IB theory for predicting probabilistic SSM from unsegmented 3D images and demonstrated that it provides better calibrated aleatoric uncertainty quantification than existing state-of-the-art techniques without sacrificing accuracy. 
By learning a latent representation in the context of the task, we provide a more scalable, flexible framework that better captures data non-linearity. 
Additionally, the proposed method is self-regularized and generalizes better under limited training data. 
These contributions increase the feasibility of using SSM in research and medicine by both bypassing the time and cost-prohibitive steps of traditional SSM and providing the necessary safeguard against model over-confidence.
This has the potential to improve medical standards and increase patient accessibility to statistic-based diagnosis. 
In future work, we plan to analyze the effectiveness of this model for predicting SSM with clinically relevant quantities, and to explore techniques for learning the optimal latent dimension size in tandem with the task. 

\vspace{-0.05in}
\subsubsection{Acknowledgements}
This work was supported by the National Institutes of Health under grant numbers NIBIB-U24EB029011, NIAMS-R01AR076120, NHLBI-R01HL135568,  NIBIB-R01EB016701, and NIGMS-P41GM103545. 
The content is solely the responsibility of the authors and does not necessarily represent the official views of the National Institutes of Health.
The authors would like to thank the University of Utah Division of Cardiovascular Medicine for providing left atrium MRI scans and segmentations from the Atrial Fibrillation projects.
\vspace{-0.05in}

%
%
%
\bibliographystyle{splncs04}
\bibliography{references}
\end{document}